\documentclass{article} 
\usepackage{iclr2022_conference,times}

\usepackage{amsmath,amsfonts,bm}

\def\eqref#1{equation~\ref{#1}}

\def\1{\bm{1}}

\DeclareMathAlphabet{\mathsfit}{\encodingdefault}{\sfdefault}{m}{sl}
\SetMathAlphabet{\mathsfit}{bold}{\encodingdefault}{\sfdefault}{bx}{n}

\usepackage{hyperref}
\usepackage{url}
\usepackage{xspace}
\usepackage{graphicx}
\usepackage{caption}
\usepackage{subcaption}
\usepackage{amsmath}%
\usepackage{MnSymbol}%
\usepackage{wasysym}%
\usepackage{lscape}

\DeclareFontEncoding{LS1}{}{}
\DeclareFontSubstitution{LS1}{stix}{m}{n}
\DeclareSymbolFont{symbols4}{LS1}{stixbb}{m}{it}
\DeclareMathSymbol{\varhexagonblack}{\mathord}{symbols4}{"DD}
\DeclareMathSymbol{\hexagonblack}   {\mathord}{symbols4}{"DE}

\newcommand{\patchmerger}{\textsc{PatchMerger}\xspace}

\newcommand{\densesym}{{\protect\scalebox{1.4}{$\bullet$}}}
\newcommand{\everysym}{{\protect\scalebox{1.0}{$\blacksquare$}}}
  
\newcommand{\vmoesym}{{\protect\scalebox{1.4}{$\mathbf{x}$}}}

\title{Learning to Merge Tokens \\ in Vision Transformers}

\iclrfinalcopy

\author{Cedric Renggli\thanks{Authors contributed equally to the work. Cedric's work was done while interning at Google Research. Correspondence to \texttt{cedric.renggli@inf.ethz.ch} and \texttt{\{andresp, rikel\}@google.com}.} \\
ETH Zurich
\And
André Susano Pinto$^*$ \\
Google Research
\And
Neil Houlsby \\
Google Research
\AND
Basil Mustafa \\
Google Research
\And
Joan Puigcerver \\
Google Research
\And
Carlos Riquelme$^*$ \\
Google Research
\AND
}

\begin{document}

\maketitle

\begin{abstract}
Transformers are widely applied to solve natural language understanding and computer vision tasks.
While scaling up these architectures leads to improved performance, it often comes at the expense of much higher computational costs.
In order for large-scale models to remain practical in real-world systems, there is a need for reducing their computational overhead.
In this work, we present the \patchmerger, a simple module that reduces the number of patches or tokens the network has to process by merging them between two consecutive intermediate layers.
We show that the \patchmerger achieves a significant speedup across various model sizes while matching the original performance both upstream and downstream after fine-tuning.
\end{abstract}
\section{Introduction}

The success of modern deep learning is tightly coupled with increasing model and dataset sizes.
This trend is especially visible in natural language processing (NLP) and in computer vision.
In NLP, some recent models such as BERT~\citep{devlin2018bert} or GPT-3~\citep{brown2020language} achieve the state-of-the-art performance on many language tasks.
One major ingredient to enable the efficient scaling of such models are transformer blocks~\citep{vaswani2017attention}.
Transformer blocks perform \textit{attention} on a token-level \citep{bahdanau2014neural}.
In language models, for instance, a token may correspond to a word.
\cite{dosovitskiy2020image} adapted this technique to the vision domain, introducing the vision transformer (ViT) architecture.
The key idea is to split the image into a sequence of square patches (e.g., $14 \times 14$ pixels) and embed them into a fixed-size representation. The patches or tokens\footnote{We use the terms \textit{token} and \textit{patch} in the context of vision transformers interchangeably to refer to the representations passed between transformer blocks.} are then fed into the transformer blocks.
\
The ViT model size and computational cost depends on (1) the number of tokens (dependent of the patch size, smaller patches means more patches in total), (2) the number of layers (i.e., how many transformer encoder blocks we apply), and (3) the dimension of the token representations (referred to as the hidden dimension).

Increasing model capacity while limiting its computational cost is an active research area.
In NLP~\citep{shazeer2017outrageously}, sparse mixtures of experts have proved successful in allowing models to scale up to trillions of parameters.
Sparse models only apply a small subset of the parameters to process each example, as dictated by trainable routers. 
This \textit{conditional computation} approach was then extended in \citet{riquelme2021scaling} to Vision Mixture of Experts (V-MoE).
Whilst V-MoE successfully increases the number of parameters by an order of magnitude, the model still handles increasingly more tokens at higher resolutions.

Further, \citet{riquelme2021scaling} showed that one can save compute by skipping the computation of expert blocks for some tokens while having very little impact on the model's performance.
They propose an algorithm to assign different priorities to each token with the hope to save compute on the least important ones. Batch Priority Routing works well even when only processing 15-30\% of the tokens through the expert blocks. Unfortunately, all tokens are still processed by the non-expert blocks (which represent a large fraction of the total blocks).


Inspired by these insights and by related work inspecting the impact of the number of tokens on the downstream accuracy  (c.f, Section~\ref{sec:background}), we explore if the number of tokens required to classify an image can be dramatically \textit{reduced} at an early stage with minimal computational extra costs while keeping the performance mostly intact.

To achieve this goal, we introduce a simple module called the \patchmerger.
This module can be placed between any two encoder blocks and it will reduce the number of tokens by merging them together.
Importantly, the number of output tokens is independent from the number of input tokens and the same network can handle a different number of input tokens. This capability is crucial to support \textit{fine-tuning} a model at different resolution than it was pre-trained with, and some naive alternative methods to alter the number of tokens fail to satisfy this requirement.

Our main contributions are as follow:
\begin{enumerate}
    \item We propose a simple, yet novel learned module called the \patchmerger to reduce the number of tokens inside Vision Transformers.
    \item We provide extensive experimental results showing that by using the \patchmerger we can achieve comparable upstream and downstream performance across all ViT and V-MoE variants, while saving a significant amount of compute and time (40-60\% for large models).
    \item We provide additional comparative studies justifying the design choices regarding where to place the \patchmerger and how aggressively to merge tokens.
\end{enumerate}
\section{Background and Related Work}
\label{sec:background}


An effective recipe to improve model performance in both NLP and computer vision is to scale the total amount of compute~\citep{brown2020language,zhai2021scaling}.
Fortunately, expensive upstream models have been shown to preserve their gains when finetuned to downstream tasks.
Consequently, researchers now compare the performance of models with a comparable overall training cost.

\cite{dosovitskiy2020image} shows that Vision Transformers can match state-of-the-art CNNs while requiring substantially fewer computational resources to train.
ViT represents an image as a sequence of patches (e.g. an image of 256x256 pixels is split into 1024 tokens of 8x8 pixels) and it processes them with the Transformer architecture \citep{vaswani2017attention}, a well-known player in the language community.
A Transformer is composed of a stack of blocks with residual connections; each of them includes a self-attention and a MLP layer.
The number of tokens is kept constant throughout the network up to a classification head.

This constant number of tokens being processed during the transformer layers is in contrast with the more typical pyramidal structure of CNN based architectures, where the input size decreases as we progress along the network.
Accordingly, there has been a number of works that play with and try to mimic this sequentially cheaper behavior, including PyramidViT~\citep{wang2021pyramid}, the Perceiver~\citep{jaegle2021perceiverIO,jaegle2021perceiverIterative} and the TokenLearner~\citep{ryoo2021tokenlearner}.
Our work is very similar in spirit to the last, whereas our implementation and merging module is conceptually simpler.
The Swin Transformer~\citep{liu2021swin} avoids quadratic cost in the number of tokens by limiting the attention range, and it merges adjacent tokens through the layers; other recent works such as the BoTNet~\citep{srinivas2021bottleneck} effectively combine convolutions and attention layers. 

Another line of work on compute efficiency is sparsity and conditional computation.
Recently, in the computer vision domain, \citet{riquelme2021scaling} have shown that by replacing the dense layer of some transformer blocks with mixture of experts one can obtain higher performance at roughly the same computational cost.
Additionally, they also show that small routing modules learnt from scratch can successfully decide when to skip the MLP processing for some tokens.
This strongly suggests that not all tokens are needed, an idea we heavily exploit in this work.
\section{The \patchmerger Module}

\begin{figure}
    \centering
    \includegraphics[width=0.95\linewidth]{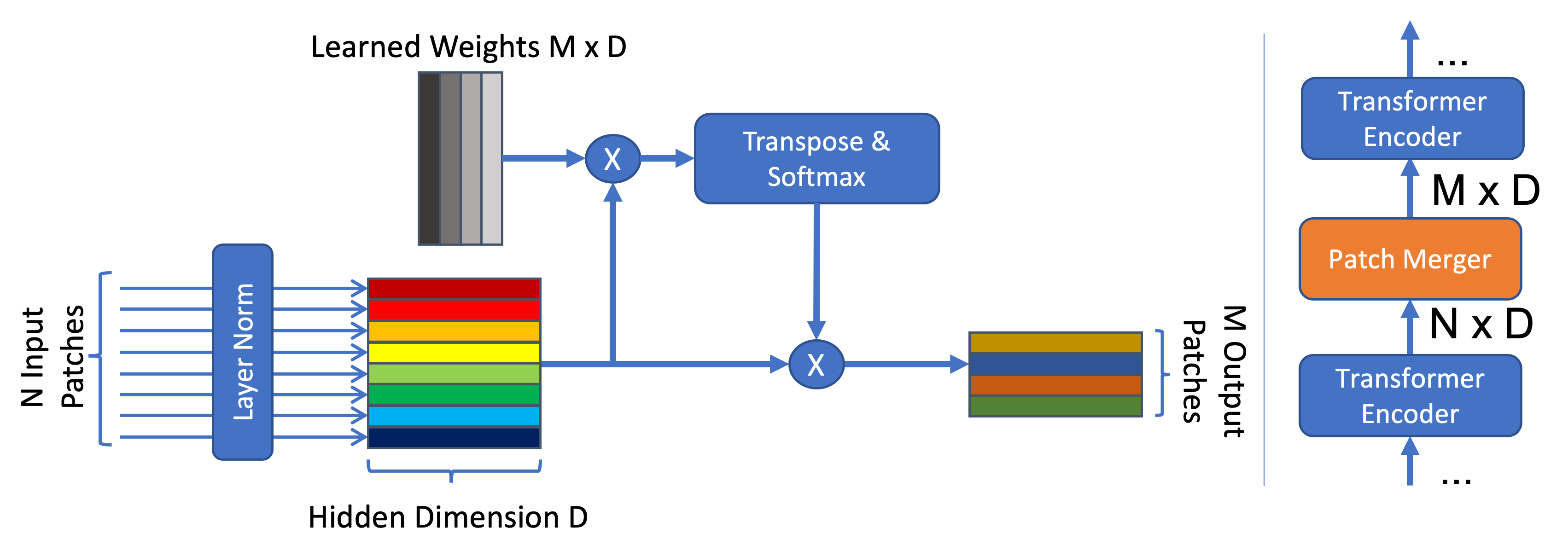}
    \caption{\textbf{(Left)} Overview of the \patchmerger module with $N$ input tokens and $M$ output ones. \\ \textbf{(Right)} Placement and token shapes when adding a \patchmerger between two encoder blocks.}
    \label{fig:illustration}
\end{figure}

The \patchmerger outputs a fixed number of tokens when applied to any number of input tokens.
We place it between two consecutive transformer encoder layers.
Accordingly, all the subsequent layers will process the new (and reduced) number of tokens.
Intuitively, the way it works is as follows.
Suppose we set the number of output patches to be $M$.
Then, for any incoming patch $p$, the \patchmerger computes $M$ scores $s_{1:M}(p)$ --- one per output patch --- based on a linear transformation of the patch representation at that point, say $z_p$.
We normalize the scores to sum up to 1 by applying a softmax transformation: $\sigma(s_{1:M}(p))$.
Finally, $p$ will contribute to each new output patch $i$ with its representation weighted by the scored computed for that output patch, that is, $\sigma(s_{i}(p)) \ z_p$.

This simple idea has several interesting implications.
First, similar patches are \emph{routed} similarly, i.e., they end up contributing or being merged into the same subset of output patches.
This can be useful, for example, to reduce the amount of compute required by redundant background patches in object-centric images.
Second, the way an old patch is distributed into new patches is independent of the patch's original \emph{position} or index, as it would happen with a raw linear merging operation.
Finally, we can freely increase or decrease the number of input patches while keeping the learned parameters intact.
If the number of training and inference tokens differ significantly, there can be a mismatch in the scale of the weighted sum for output patches.
However, a layer norm is applied right after the \patchmerger, in the following transformer block, which helps to mitigate this potential effect.
Moreover, our experiments suggest finetuning is able to cope and adapt to this extremely quickly.

\paragraph{Connections to Attention.}
Figure~\ref{fig:illustration} gives a graphical overview of the \patchmerger.
Layer normalization is applied before merging blocks.
The \patchmerger then consists of a learned matrix $W \in \mathbb{R}^{D \times M}$, where $D$ is the size of the latent token representation.
If we represent the $N$ input patches after layer normalization by a matrix $\mathbf{X} \in \mathbb{R}^{N \times D}$, the merged representations $\mathbf{Y} \in \mathbb{R}^{M \times D}$ are computed as follows:
\begin{equation}
\label{eq:patchmergers}
    \mathbf{Y} = \textrm{softmax}\left(\left(\mathbf{X}W\right)^T\right) \mathbf{X} = \textrm{softmax}\left(W^T\mathbf{X}^T\right) \mathbf{X}.
\end{equation}
Equation (\ref{eq:patchmergers}) is similar to the non-scaled dot-product attention formula from~\citet{vaswani2017attention}:
\begin{equation}\label{eq:dot_product_attention}
\text{Att}(Q, K, V) = \text{softmax}\left(QK^T\right) V.
\end{equation}
In particular, we use the input patches as both keys and values, and simply learn a fixed set of queries --- one per output patch. This is partially similar to the default implementation of encoder-decoder attention, where both keys and values are linear projections of the encoder output~\citep{bahdanau2014neural}.
However, their queries are computed based on the output of the previous decoder layer.
Other recent works have also explored input-independent learnable queries~\citep{jaegle2021perceiverIO}.
Moreover, unlike the dot-product attention in (\ref{eq:dot_product_attention}), the softmax in (\ref{eq:patchmergers}) is applied per input token (i.e. along each column), typically referred to as \emph{bottom-up} attention~\citep{abnar2019attn}.
We tried to apply this idea directly to the attention layer in one of the transformer blocks, but it led to severe instabilities (and we had to deal with the mismatch in shapes due to the residual connection).
Consequently, we decided to just keep \patchmerger modules in-between full transformer blocks.
\section{Experimental Evaluation}
In this section we present experiments showing that using the \patchmerger matches the performance of its backbone network (either VIT or V-MoE) while requiring significant fewer FLOPs and runtime (between 40\%-50\% for the largest models).

\textbf{Training Data.}
We train all of our models on JFT~\citep{hinton2015distilling}, a proprietary dataset containing ~300M images. Each image is classified to one or several of the 18291 classes, mostly natural images (animals, landscapes, objects, etc), and we use a (224 $\times$ 224 $\times$ 3) resolution.
We follow the model sizes and training strategies from~\citep{dosovitskiy2020image,riquelme2021scaling}: training for 5 epochs for Small, 7 for Base, and 14 for the Large and Huge models.
The patch sizes are $(32, 32)$ for Small and Base, $(16, 16)$ for Large, and $(14, 14)$ for Huge, leading to 49, 196, and 256 training patches per image respectively (note we usually append an additional cls token).
We report the precision at one in a test set, that is, how often the top-1 model prediction is indeed one of the possibly multiple labels for the test image.

\textbf{Transfer to ImageNet.}
We want to train \emph{transferable} models, so that an expensive training can be amortized over a wide range of downstream tasks.
We test this property in two ways: via few-shot evaluation and full finetuning.
In particular, we transfer to ImageNet~\citep{deng2009imagenet}.
For few-shot evaluation, we freeze the model weights, and replace the original JFT head with a new from-scratch head for ImageNet.
Then, we train only the head on a few examples per class (5-shot corresponds to 5 examples per ImageNet class, and so on).
For fine-tuning, we train all the JFT-checkpoint weights on the whole ImageNet dataset for a few epochs.
We report the top-1 model accuracy.

\textbf{PatchMerger Design Choices.}
As introduced in the previous section, we can use the \patchmerger in several ways.
We experimented with many design choices --a single or several mergers, more or less aggressive token reduction, etc.--, while in this section we only focus on the ones that proved simplest and most successful.
Given a backbone network, we place one merger layer in the middle of the network, and it outputs 8 patches.
All the backbone networks have an even number of encoder blocks (that is, 8, 12, 24, or 32), and we place the merger between the first and second half of blocks (i.e., after the first 4, 6, 12, or 16 blocks).
The number of tokens that backbone networks use depends on the patch size; however, regardless of the number of initial patches, we always merge them into 8 patches.
This approach dramatically reduces the amount of compute required in the second half of encoder blocks, whereas the amount of compute in the first half does not change.
Later in this section we independently test each of these design choices (where to place the merger, and how much to merge).

A common practice when finetuning a pre-trained model consists in increasing the image size for the new task.
If the patch size remains constant, then the pre-trained embedding layer can be loaded, applied, and finetuned as usual.
As images are larger, the total number of patches then increases.
However, the \patchmerger does \emph{not} adjust the bottleneck: we still merge in the middle of the network all the incoming patches into just 8 of them.
This means the amount of downstream processing will grow with respect to the upstream (or pre-trained) model in the first half of the network, but the second half's compute will remain intact after the \patchmerger.
For the original backbone network (i.e., without \patchmerger module), however, compute grows everywhere.
Thus, the compute gap between the \patchmerger equipped network and its backbone will be even more severe at finetuning and inference time.

\begin{figure}
\centering
\begin{subfigure}{.45\textwidth}
  \centering
  \includegraphics[width=\linewidth]{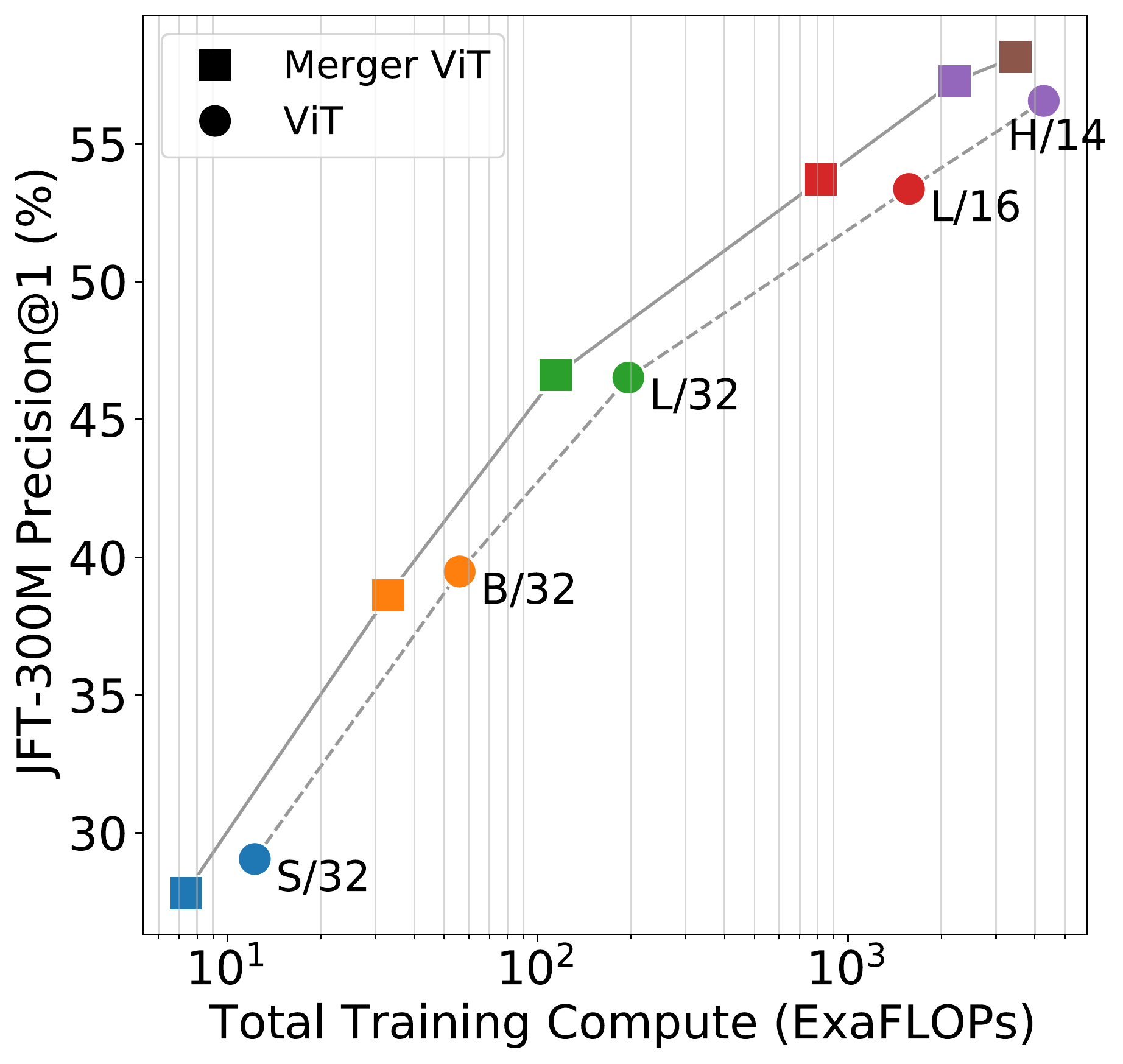}
  \caption{Upstream precision at one.}
  \label{fig:vit_upstream_fewshow_sub1}
\end{subfigure}%
\begin{subfigure}{.45\textwidth}
  \centering
  \includegraphics[width=\linewidth]{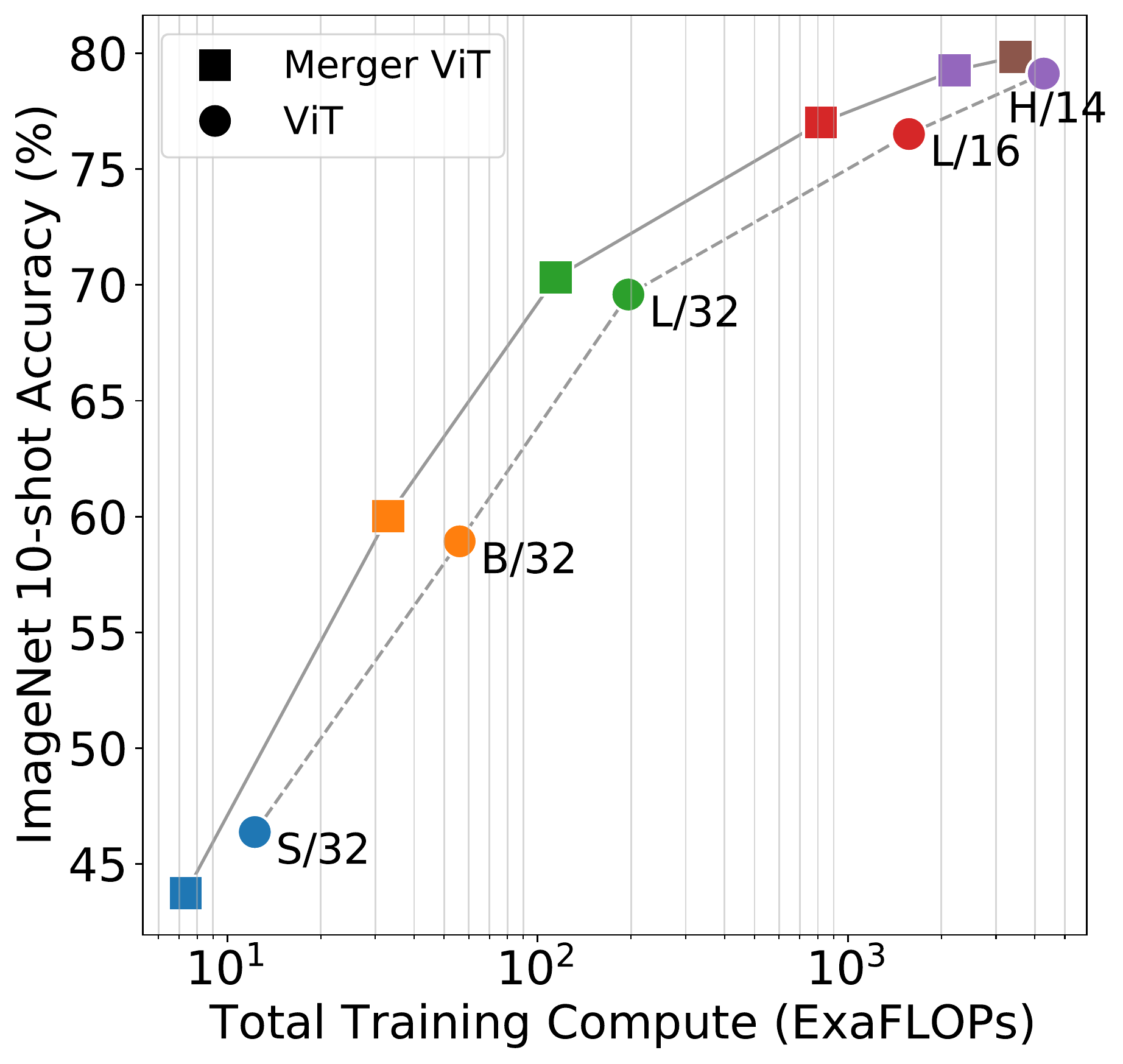}
  \caption{10-shot accuracy on ImageNet.}
  \label{fig:vit_upstream_fewshow_sub2}
\end{subfigure}
\caption{
\textbf{Patch Merger for ViT}. Performance versus total training FLOPs. Colors represent different ViT variants, markers represent either standard $\densesym{}$ ViT or $\everysym{}$ Merger ViT. The brown Merger ViT (top-right most) corresponds to H/11. %
Lines show the Pareto frontier of Merger ViT (solid) and ViT (dashed).
Figure~\ref{fig:vit_upstream_fewshow_runtime} in the Appendix shows performance versus runtime for these models.
\label{fig:vit_upstream_fewshow}
}
\end{figure}

\begin{figure}
\centering
\begin{subfigure}{.45\textwidth}
  \centering
  \includegraphics[width=\linewidth]{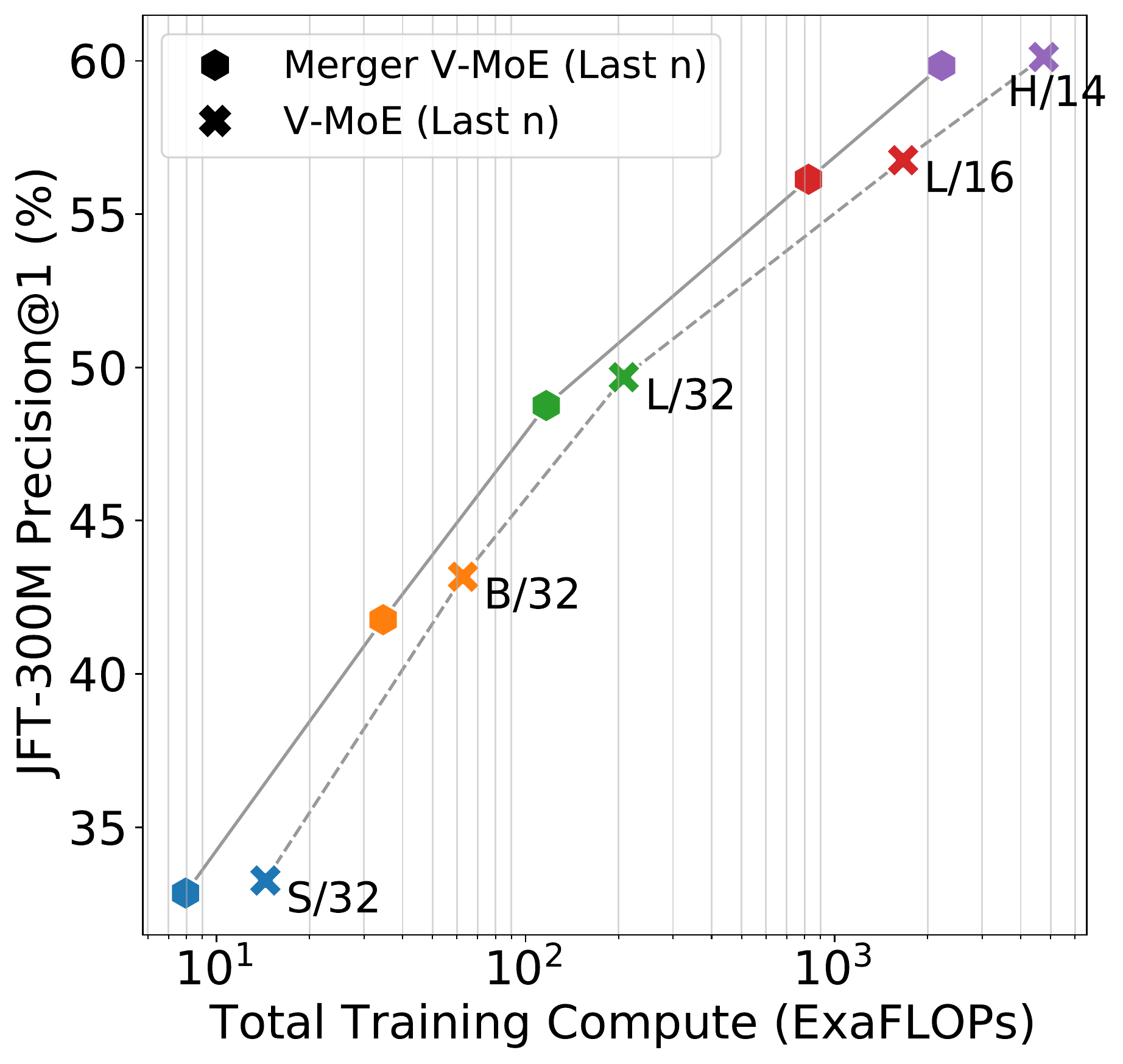}
  \caption{Upstream precision at one.}
  \label{fig:vmoe_upstream_fewshow_sub1}
\end{subfigure}%
\begin{subfigure}{.45\textwidth}
  \centering
  \includegraphics[width=\linewidth]{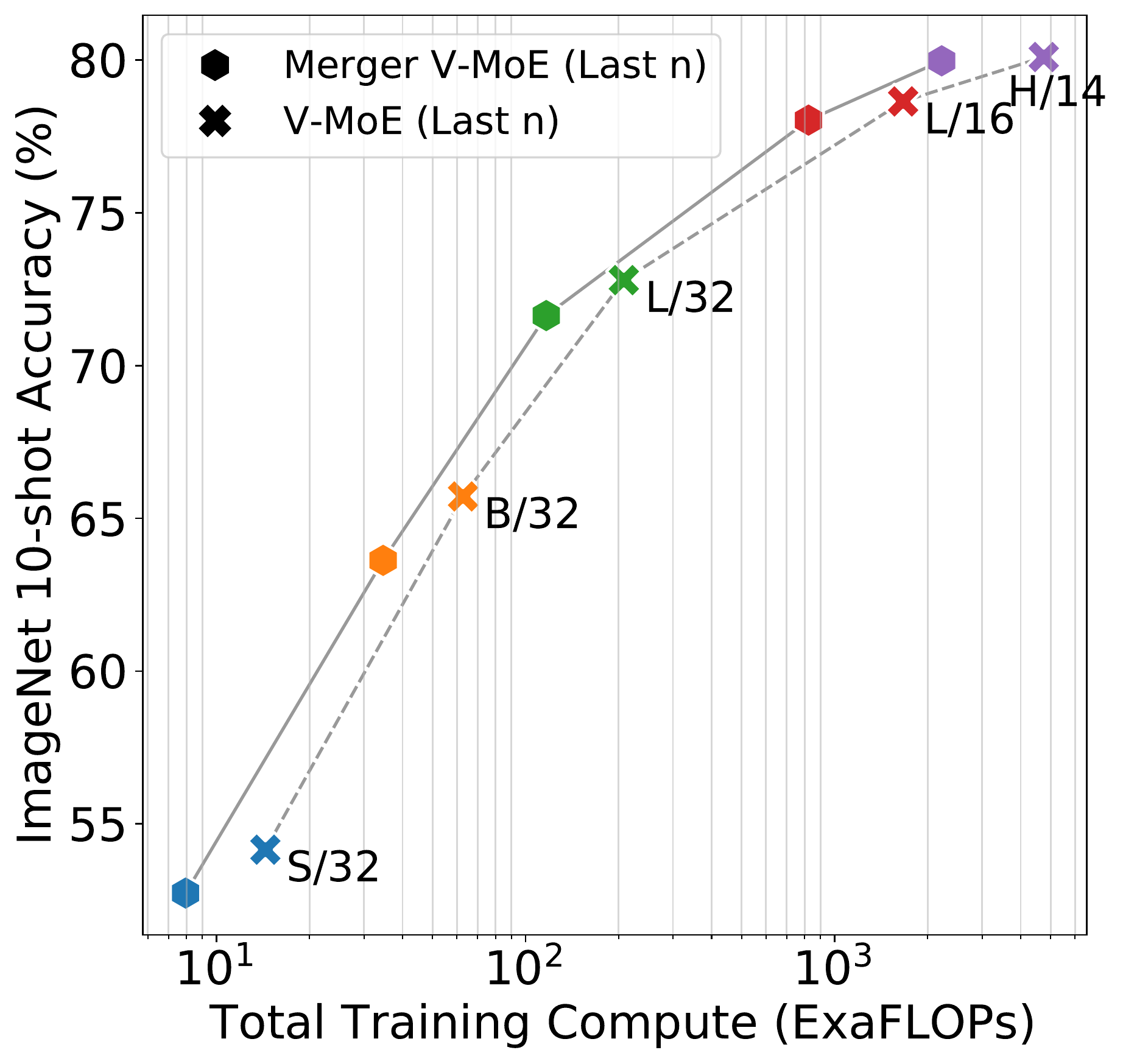}
  \caption{10-shot accuracy on ImageNet.}
  \label{fig:vmoe_upstream_fewshow_sub2}
\end{subfigure}
\caption{
\textbf{Patch Merger for V-MoE}. Performance versus total training FLOPs. Colors represent different V-MoE variants, markers represent either standard $\vmoesym{}$ V-MoE or $\varhexagonblack$ Merger V-MoE.
Lines show the Pareto frontier of Merger V-MoE (solid) and V-MoE (dashed).
Figure~\ref{fig:vmoe_upstream_fewshow_runtime} in the Appendix shows performance versus runtime for these models.
\label{fig:vmoe_upstream_fewshow}}
\end{figure}

\textbf{Results.}
We first present upstream and few-shot results.
Figure~\ref{fig:vit_upstream_fewshow} and Figure~\ref{fig:vmoe_upstream_fewshow} show the performance versus training FLOPs trade-offs for ViT and V-MoE respectively.
For completeness, Figure~\ref{fig:vit_upstream_fewshow_runtime} and Figure~\ref{fig:vmoe_upstream_fewshow_runtime} in the appendix present our results with respect to training runtime (rather than FLOPs).
All numbers are presented in Table~\ref{table:vit} (ViT) and Table~\ref{table:vmoe} (V-MoE).

\begin{figure}
\centering
\begin{subfigure}{.45\textwidth}
  \centering
  \includegraphics[width=\linewidth]{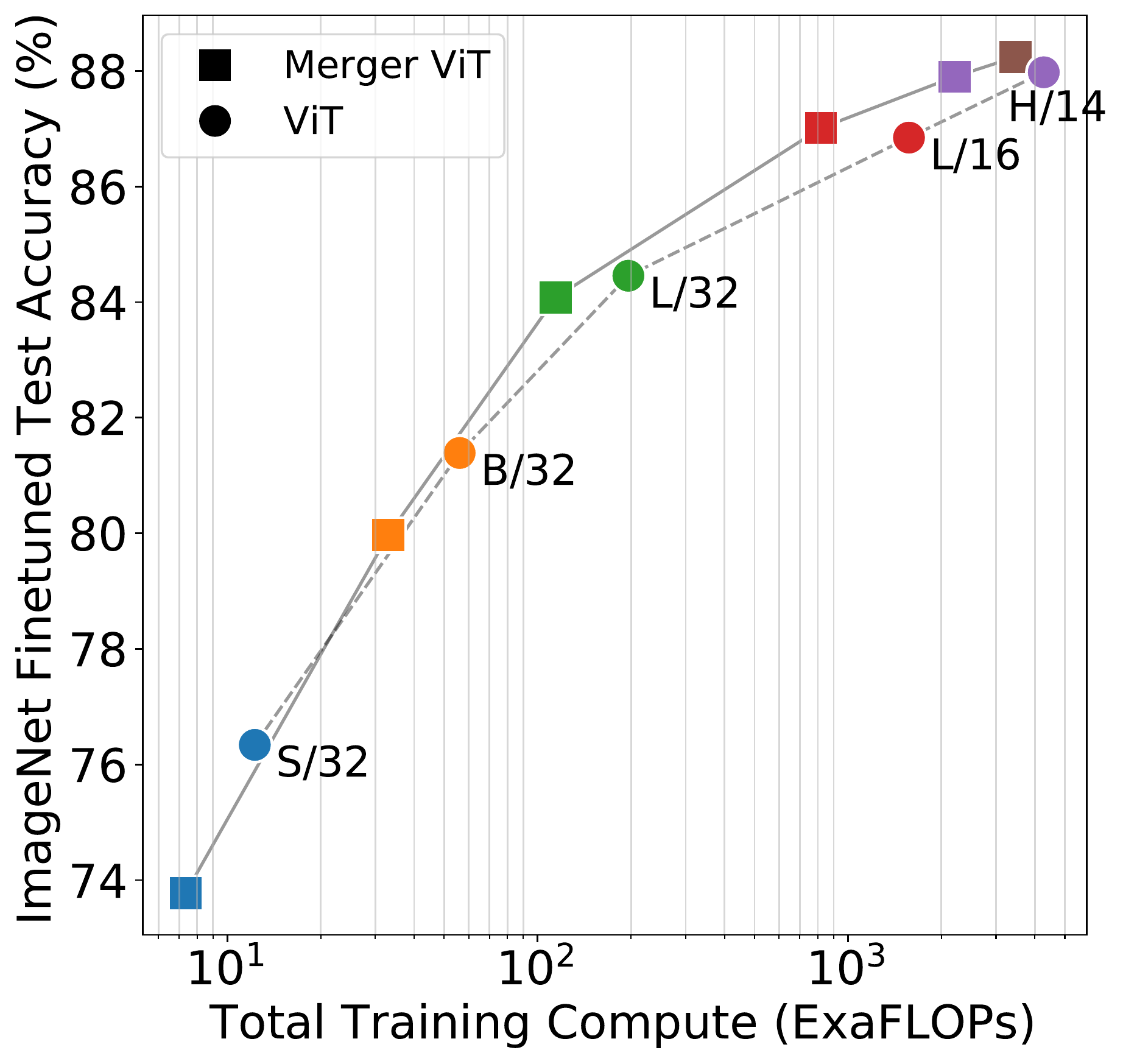}
  \caption{\textbf{ViT.} Finetuning accuracy on Imagenet.}
  \label{fig:downstream_sub1}
\end{subfigure}%
\begin{subfigure}{.45\textwidth}
  \centering
  \includegraphics[width=\linewidth]{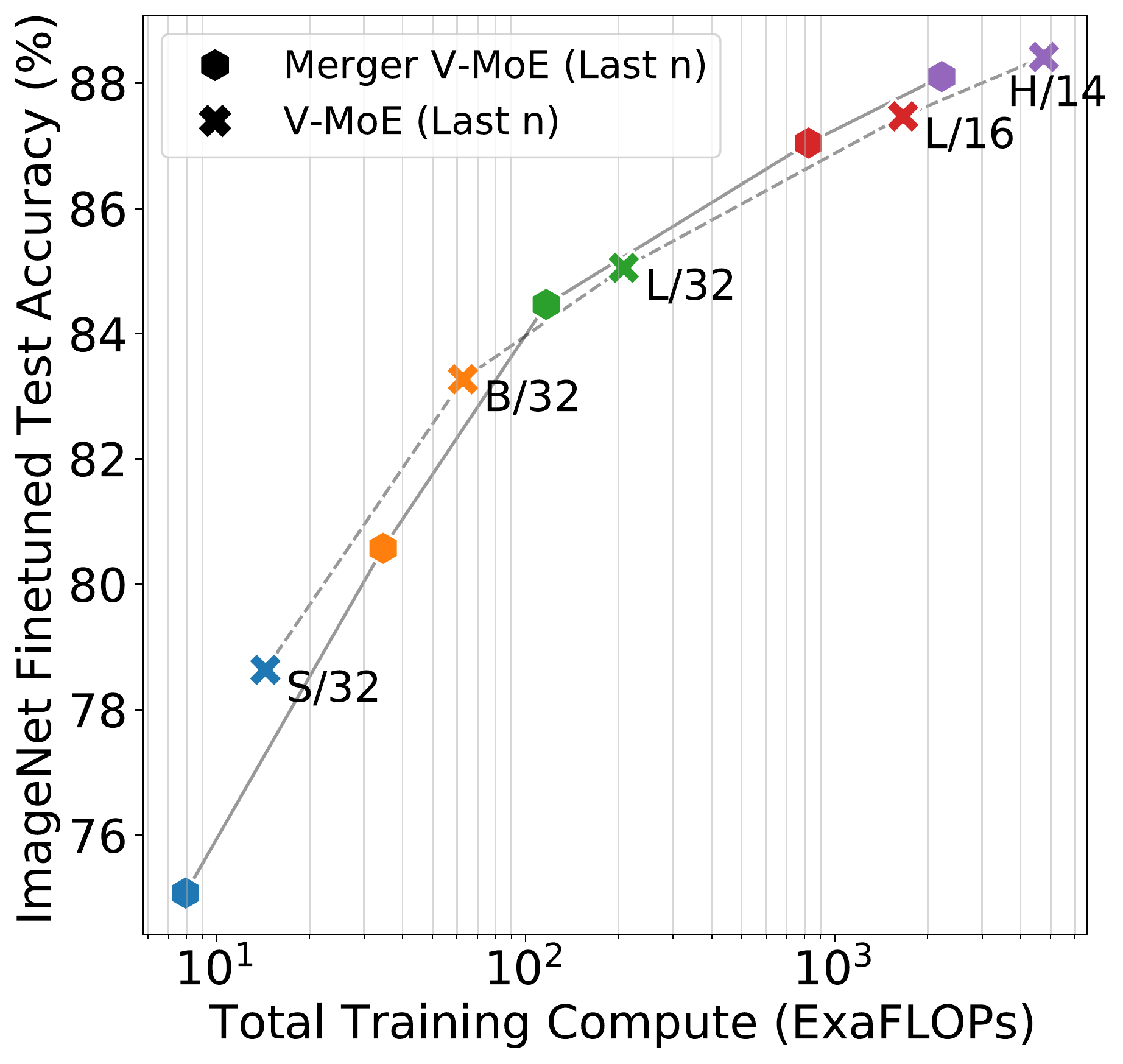}
  \caption{\textbf{V-MoE.} Finetuning accuracy on Imagenet.}
  \label{fig:downstream_sub2}
\end{subfigure}
\caption{
\textbf{Full finetuning on ImageNet}. Performance versus total training FLOPs. Colors represent different ViT and V-MoE variants (including H/11 for Merger ViT), markers represent either standard $\densesym{}$ ViT, $\everysym{}$ Merger ViT, $\vmoesym{}$ V-MoE or $\varhexagonblack$ Merger V-MoE.
Lines show the Pareto frontiers in each case.
Figure~\ref{fig:downstream_runtime} in the Appendix shows performance versus runtime for these models.
\label{fig:downstream}}
\end{figure}

The Merger ViT (Figure~\ref{fig:vit_upstream_fewshow}) obtains comparable --sometimes even better-- performance than its backbone ViT models with a much lower cost.
In particular, the Pareto frontier for Merger ViT models completely dominates the ViT one, and by a wide margin.
When fully fine-tuning on ImageNet, shown in Figure~\ref{fig:downstream}a), we also see that the large merged models (L/32, L/16, and H/14) match their backbone counterparts.

Moreover, at the largest scale, the Merger ViT-H/14 reaches the same performance as ViT-H/14 both in JFT and ImageNet 10-shot while requiring only 51.6\% of the FLOPs and 63.5\% of the runtime.
There are several ways to use these savings in compute to improve the model even further; for instance, by adding more transformer blocks or by using more tokens to start with.
In Figure~\ref{fig:vit_upstream_fewshow} we show the latter approach: Merger ViT-H/11 where we use the VIT-H backbone network (32 encoder blocks) but with a smaller patch size ($11 \times 11$).
This implies more tokens in the first half of the network, whereas the second half requires the same compute as only 8 tokens travel through it.
As shown in Figure~\ref{fig:vit_upstream_fewshow} and Figure~\ref{fig:downstream}, while this approach is still significantly cheaper than the standard VIT-H/14 (takes $\sim80\%$ of the FLOPs and runtime), it definitely outperforms it both upstream, at few-shot, and after full finetuning.

When applied to MoE models (Figure~\ref{fig:vmoe_upstream_fewshow}), we reach similar positive conclusions for large models (L/16 and H/14), where savings amount to around 53\% of the FLOPs and 44\% of the runtime.
Again, the Pareto frontier for merged MoE models dominates that of their MoE backbone, whereas we notice slight losses in performance for the small experts models.
Figure~\ref{fig:downstream}b) shows that small merged expert models underperform their backbones after finetuning too, while being much cheaper.
For large models, the finetuning Pareto frontier for merged models seems to offer a modest improvement.
Note that we add expert layers only in the last-2 non-consecutive encoder blocks (last-5 for H/14) as proposed in \citet{riquelme2021scaling}.
As the MoEs are placed in the second half of the network, all expert layers and their routers only receive 8 tokens per input image, which can decrease the expert utilization rate as compared to the case with more tokens per image.

\subsection{Comparative Studies}
In this subsection we briefly explore the two main algorithmic design choices for the \patchmerger, namely, where to place the merger and how many tokens to output.
Obviously, the later we place it, and the more tokens we output, the better performance we expect --as the model would be closer to the original backbone.
However, we also want to drastically reduce the training cost, which requires the opposite.

\textbf{Where to place the \patchmerger.}
While one could place several mergers along the network and sequentially reduce the total number of tokens, for the sake of simplicity we constrain our experiments to a single merger, and try to figure out the performance trade-offs offered by placing the merger before each encoder block.

\begin{figure}
    \centering
    \includegraphics[width=0.85\linewidth]{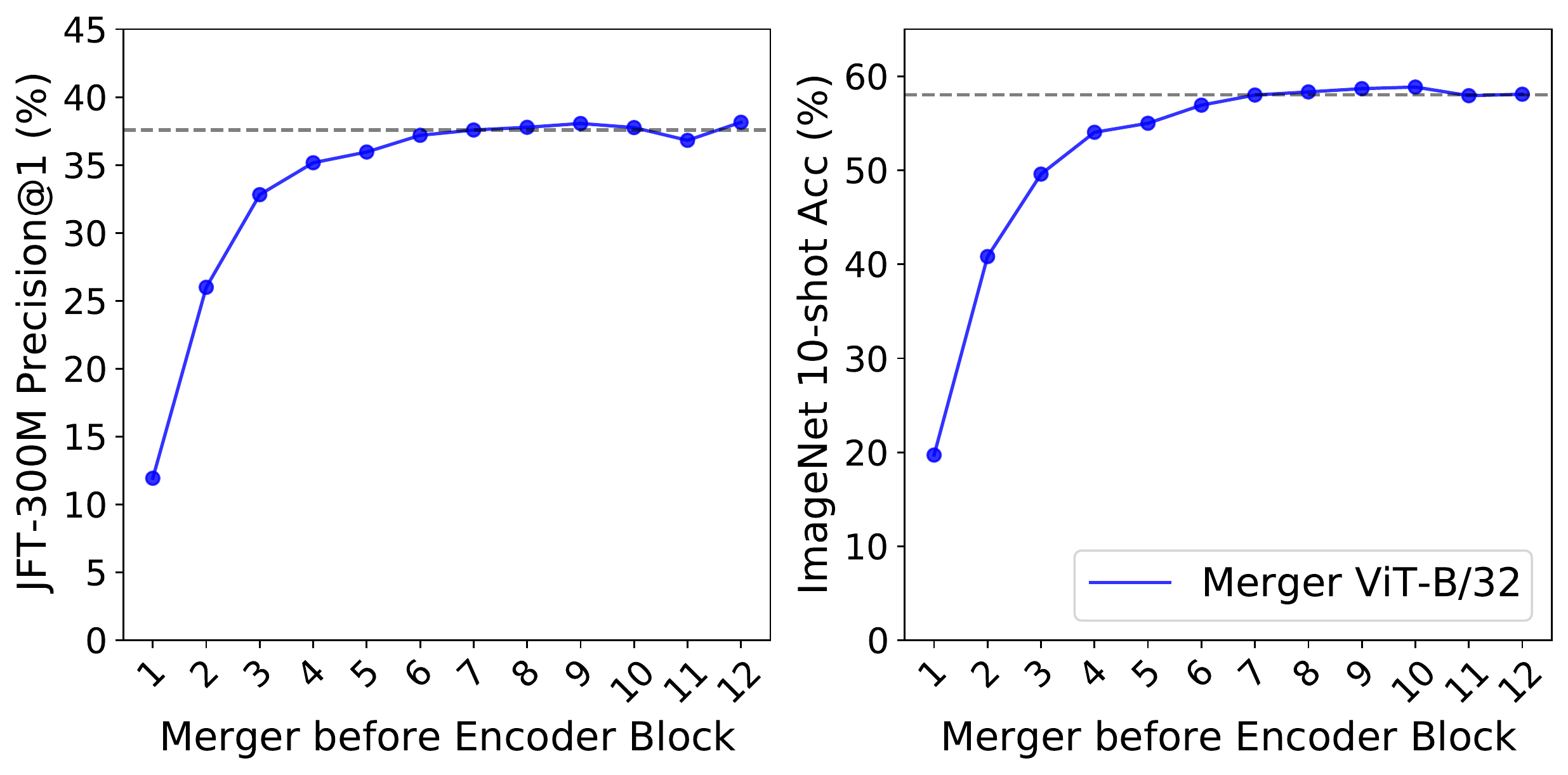}
    \caption{\textbf{Placement Comparison.} The discontinuous line corresponds to placing the merger in the middle of the network as done in previous sections --after the 6th, before the 7th encoder block, in this case.
    The number of output tokens is 8 in all cases.
    }
    \label{fig:merger_placement}
\end{figure}

Figure~\ref{fig:merger_placement} shows one such experiment with a ViT-B/32 model with 12 encoder blocks.
We show the upstream precision@1 on JFT, and the ImageNet 10/shot performance.
For each encoder block, we show the performance of a model trained with a merger placed right \emph{before} the encoder block.
We see that placing the merger too early in the network --while leading to extremely cheap models-- definitely hurts performance, as the model still needs to use more tokens.
On the other hand, around half of the network --before the sixth block--, it seems that the merged models achieve full and comparable performance.
This experiment justifies our choice in the previous section.

\begin{figure}
    \centering
    \includegraphics[width=0.85\linewidth]{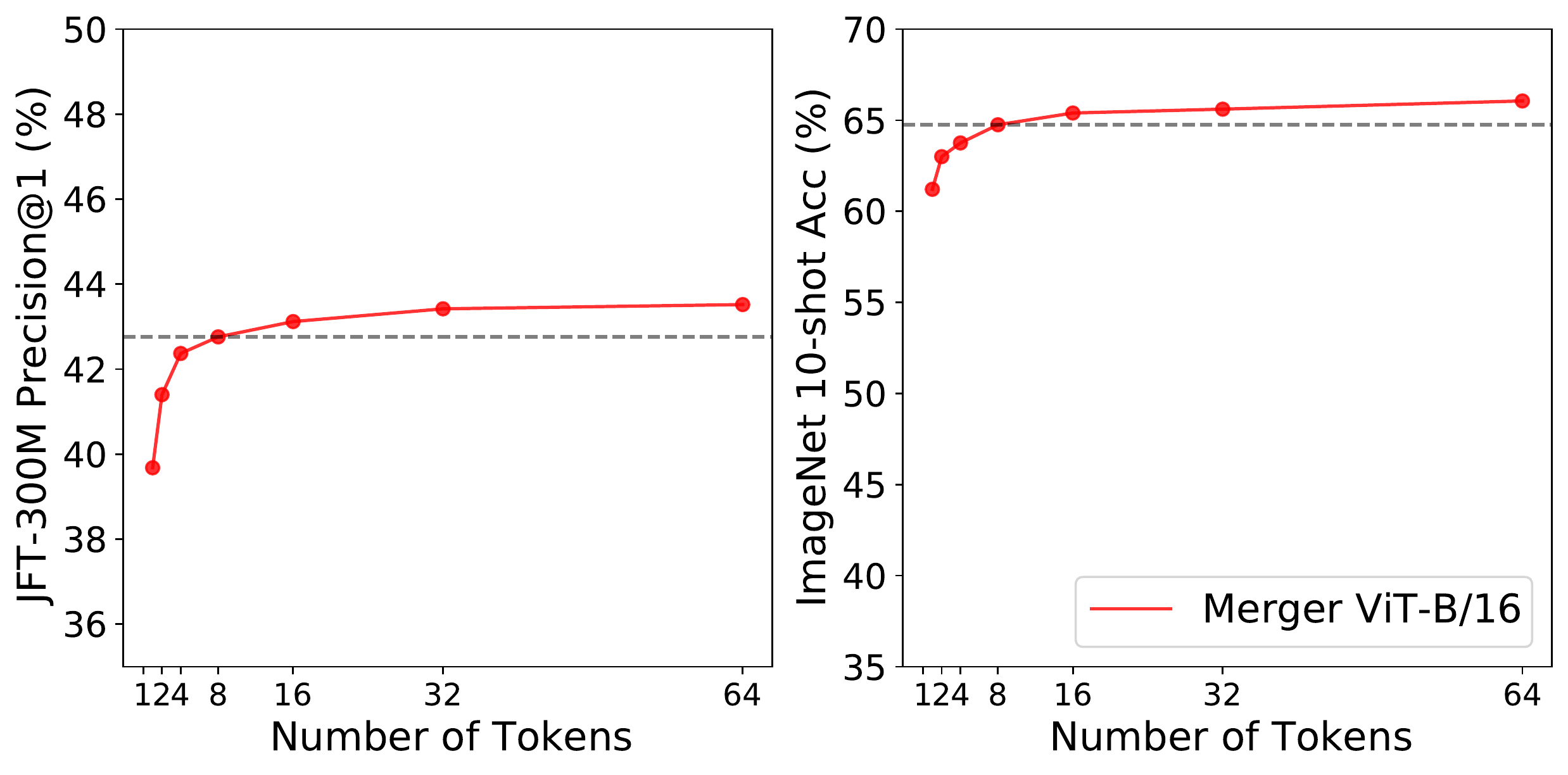}
    \caption{\textbf{Number of Output Tokens Comparison.} The discontinuous line corresponds to outputting 8 tokens as done in previous sections.
    The merger is placed in the middle of the network (after the 6th, before the 7-th encoder block), in all cases.}
    \label{fig:merger_tokens}
\end{figure}

\textbf{How many tokens to output.}
Figure~\ref{fig:merger_tokens} addresses another key question: what is the right number of tokens to keep after merging.
While the answer can be model and input patch-size dependent, we experiment with a ViT-B/16 model in this case --as it receives more tokens than a ViT-B/32 and could offer a better intuition for larger models.
As we expected, keeping more tokens is better.
On one hand, if we keep as few as 1 or 2 tokens only, the impact on performance is severe.
At the other extreme, it seems that keeping beyond 32 tokens of the 196 input tokens only offers diminishing returns.
Accordingly, we believe that a number of tokens in the range of 8-32 could be optimal.
These are ViT models; however, for expert models having a few more tokens per input can make routing and balance easier at the expense of slightly higher computational cost.


\section{Conclusion and Future Work}
We have shown that a simple module can effectively merge and combine together tokens for ViT and V-MoE, thus leading to much cheaper and practical models while preserving the original performance in most cases.
In light of the recent trends of scaling models to massive sizes, we believe this type of research will have a significant practical and environmental impact.
The \patchmerger assigns a fixed (and reduced) amount of compute to all inputs; however, not all inputs may require the same amount of processing.
Exploring mechanisms that reduce the overall training cost by assigning adaptive per-input compute levels is an exciting avenue of future research.

\bibliography{main}
\bibliographystyle{iclr2022_conference}

\appendix

\section{Appendix}
\label{sec:appendix}

\begin{table}[h!]
\begin{center}
\hspace*{-2cm}
\begin{tabular}{||c | c c c c c c c||} 
 \hline
 Model & TPU-Days & ExoFLOPs & JFT Prec@1 & INet/1shot & INet/5shot & INet/10shot & INet/finetuned  \\ [0.5ex] 
 \hline\hline
ViT-H/14  & 2387.71 & 4275.92 & 56.56 & 61.35 & 76.76 & 79.12 & 87.97 \\
\hline
Merger ViT-H/11   & 1920.82 & 3464.17 & 58.15 & 62.77 & 77.53 & 79.84 & 88.24 \\
\hline
Merger ViT-H/14   & 1518.49 & 2207.33 & 57.27 & 62.07 & 77.25 & 79.26 & 87.90 \\
\hline
ViT-L/16  & 651.18 & 1572.74 & 53.37 & 59.04 & 73.92 & 76.51 & 86.84 \\
\hline
Merger ViT-L/16   & 388.44 & 819.29 & 53.71 & 59.99 & 74.99 & 77.01 & 87.01 \\
\hline
ViT-L/32  & 97.29 & 196.11 & 46.52 & 50.34 & 66.59 & 69.58 & 84.45 \\
\hline
Merger ViT-L/32   & 66.49 & 114.38 & 46.61 & 51.29 & 67.61 & 70.32 & 84.08 \\
\hline
ViT-B/16  & 95.04 & 224.45 & 44.58 & 48.21 & 63.50 & 66.94 & -- \\
\hline
Merger ViT-B/16   & 54.22 & 117.22 & 43.75 & 49.49 & 64.35 & 67.15 & -- \\
\hline
ViT-B/32  & 27.62 & 56.07 & 39.48 & 40.66 & 55.47 & 58.93 & 81.39 \\
\hline
Merger ViT-B/32   & 17.76 & 32.99 & 38.62 & 40.53 & 57.23 & 60.02 & 79.96 \\
\hline
ViT-S/32  & 7.22 & 12.27 & 29.05 & 29.37 & 43.21 & 46.38 & 76.34 \\
\hline
Merger ViT-S/32   & 4.83 & 7.34 & 27.82 & 26.42 & 40.27 & 43.75 & 73.77 \\
 \hline
\end{tabular}
\end{center}
\caption{ViT and Merged ViT Models. TPU-days and FLOPs refer to the total training cost.}
\label{table:vit}
\end{table}
\begin{table}[ht!]
\begin{center}
\hspace*{-2cm}
\begin{tabular}{||c | c c c c c c c||} 
 \hline
 Model & TPU-Days & ExoFLOPs & JFT Prec@1 & INet/1shot & INet/5shot & INet/10shot & INet/finetuned  \\ [0.5ex] 
 \hline\hline
V-MoE-H/14    & 2735.70 & 4750.73 & 60.12 & 62.95 & 78.08 & 80.10 & 88.41 \\
\hline
Merger V-MoE-H/14     & 1520.59 & 2220.96 & 59.84 & 63.21 & 77.56 & 79.97 & 88.10 \\
\hline
V-MoE-L/16    & 761.27 & 1666.10 & 56.76 & 61.46 & 76.53 & 78.64 & 87.47 \\
\hline
Merger V-MoE-L/16     & 388.19 & 823.54 & 56.14 & 61.72 & 75.86 & 78.03 & 87.04 \\
\hline
V-MoE-L/32    & 110.65 & 207.94 & 49.68 & 54.52 & 69.90 & 72.80 & 85.05 \\
\hline
Merger V-MoE-L/32     & 67.11 & 116.70 & 48.75 & 52.03 & 68.98 & 71.63 & 84.46 \\
\hline
V-MoE-B/16    & 130.86 & 250.70 & 48.31 & 54.92 & 68.84 & 71.81 & -- \\
\hline
Merger V-MoE-B/16     & 54.78 & 118.59 & 46.65 & 53.33 & 67.53 & 70.05 & -- \\
\hline
V-MoE-B/32    & 36.80 & 62.75 & 43.17 & 48.04 & 62.45 & 65.72 & 83.27 \\
\hline 
Merger V-MoE-B/32     & 18.51 & 34.61 & 41.77 & 44.88 & 60.24 & 63.62 & 80.57 \\
\hline
V-MoE-S/32    & 12.40 & 14.40 & 33.26 & 35.49 & 50.90 & 54.16 & 78.63 \\
\hline
Merger V-MoE-S/32     & 5.33 & 7.95 & 32.85 & 33.59 & 49.23 & 52.74 & 75.07 \\
\hline
\end{tabular}
\end{center}
\caption{V-MoE and Merged V-MoE Models. TPU-days and FLOPs refer to the total training cost.}
\label{table:vmoe}
\end{table}

\clearpage

\begin{figure}
\centering
\begin{subfigure}{.45\textwidth}
  \centering
  \includegraphics[width=\linewidth]{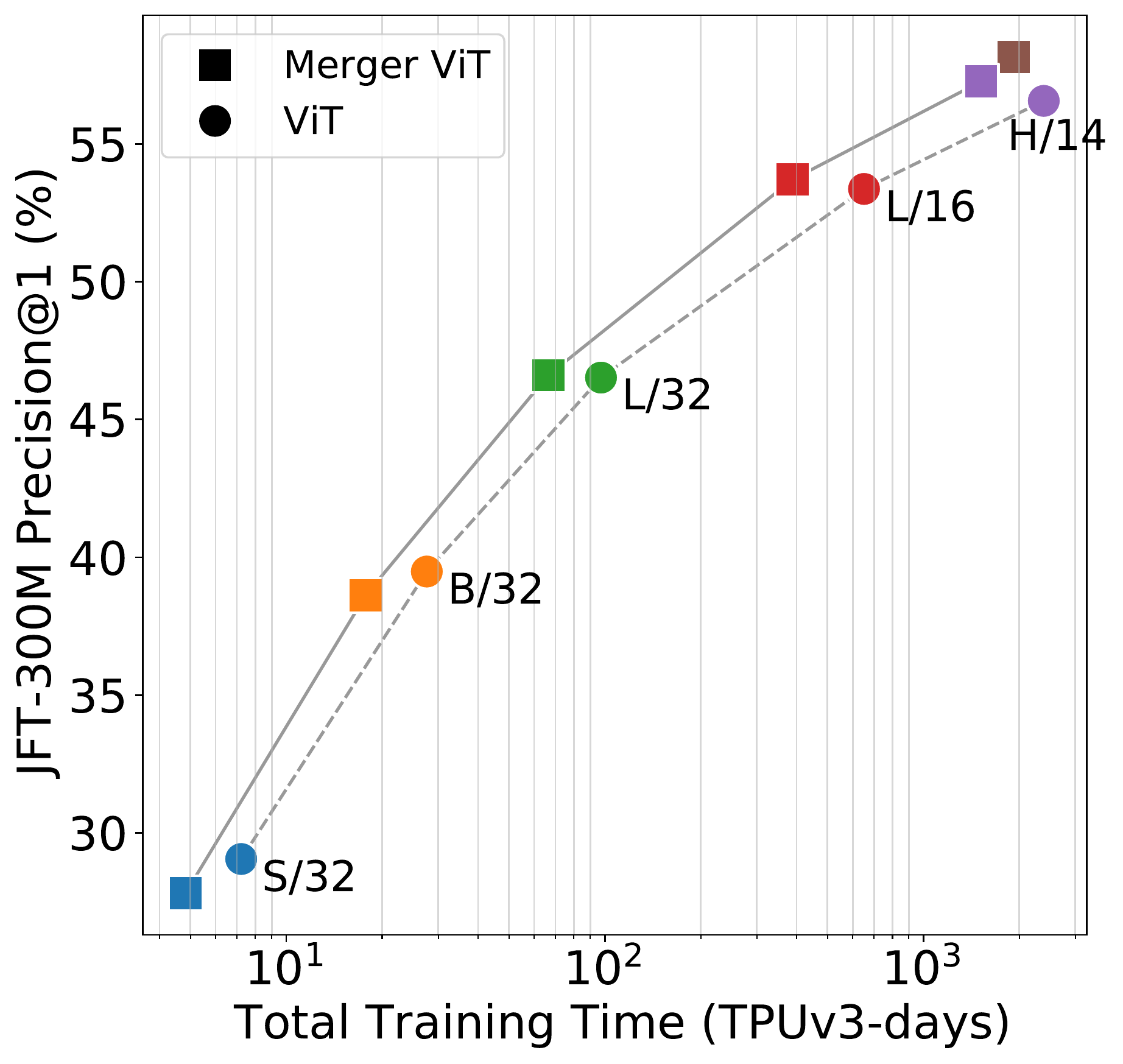}
  \caption{Upstream precision at one.}
  \label{fig:vit_upstream_fewshow_runtime_sub1}
\end{subfigure}%
\begin{subfigure}{.45\textwidth}
  \centering
  \includegraphics[width=\linewidth]{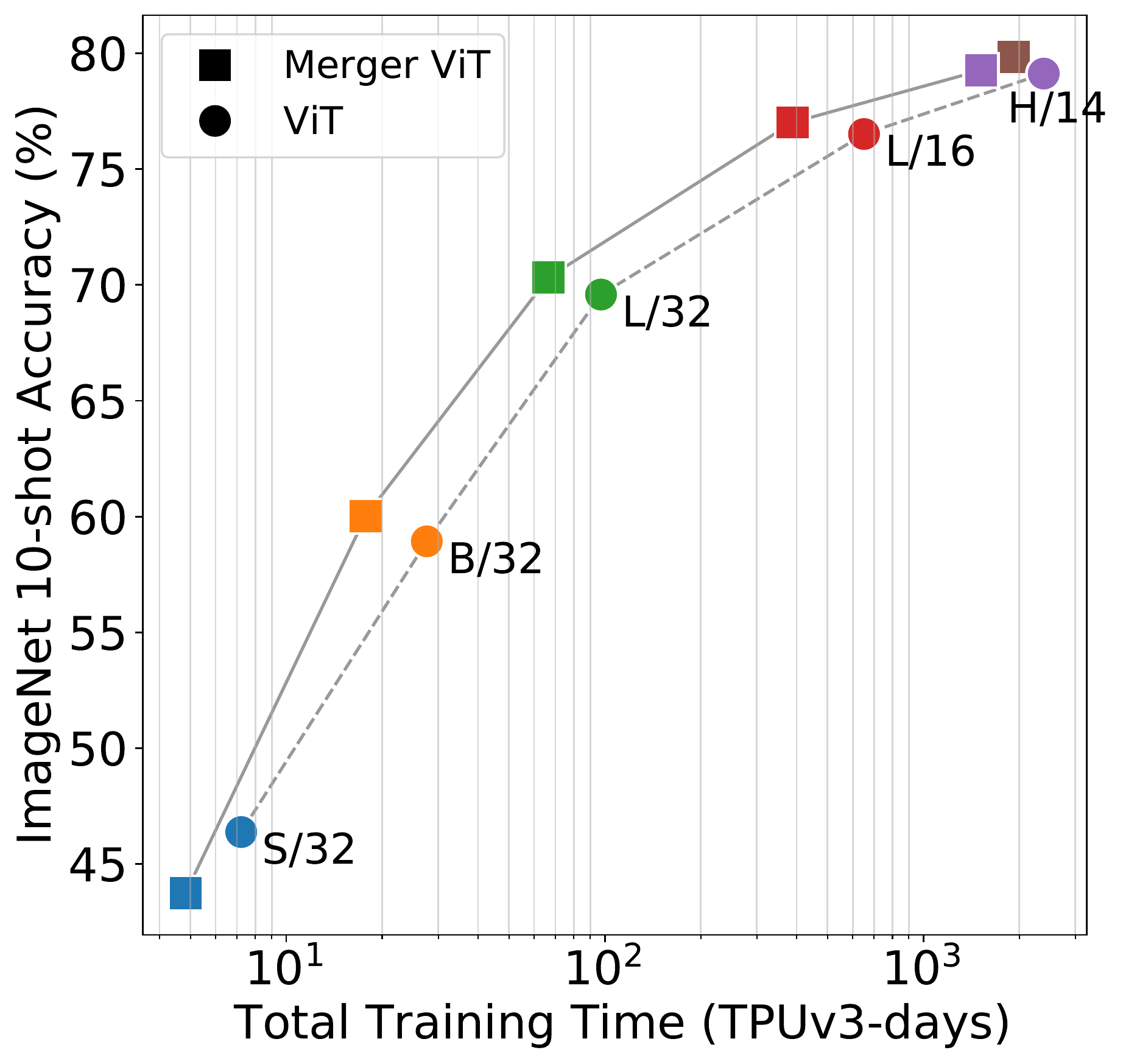}
  \caption{10-shot accuracy on ImageNet.}
  \label{fig:vit_upstream_fewshow_runtime_sub2}
\end{subfigure}
\caption{
\textbf{Patch Merger for ViT}. Performance versus total training runtime. Colors represent different ViT variants, markers represent either standard $\densesym{}$ ViT or $\everysym{}$ Merger ViT. The brown Merger ViT (top-right most) corresponds to H/11. %
Lines show the Pareto frontier of Merger ViT (solid) and ViT (dashed).
Figure~\ref{fig:vit_upstream_fewshow} in the main text shows performance versus FLOPs for these models.
\label{fig:vit_upstream_fewshow_runtime}
}
\end{figure}

\begin{figure}
\centering
\begin{subfigure}{.45\textwidth}
  \centering
  \includegraphics[width=\linewidth]{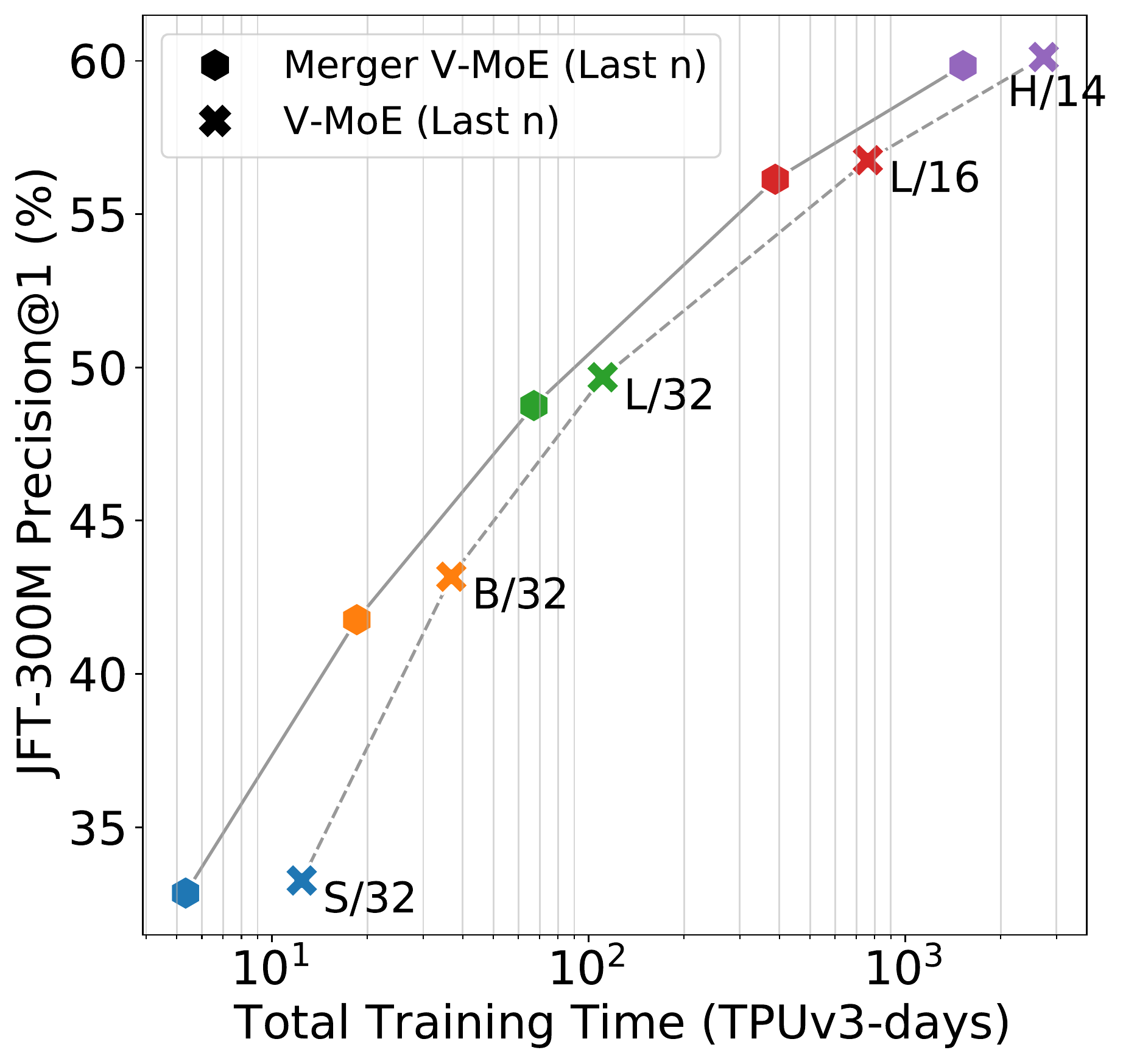}
  \caption{Upstream precision at one.}
  \label{fig:vmoe_upstream_fewshow_runtime_sub1}
\end{subfigure}%
\begin{subfigure}{.45\textwidth}
  \centering
  \includegraphics[width=\linewidth]{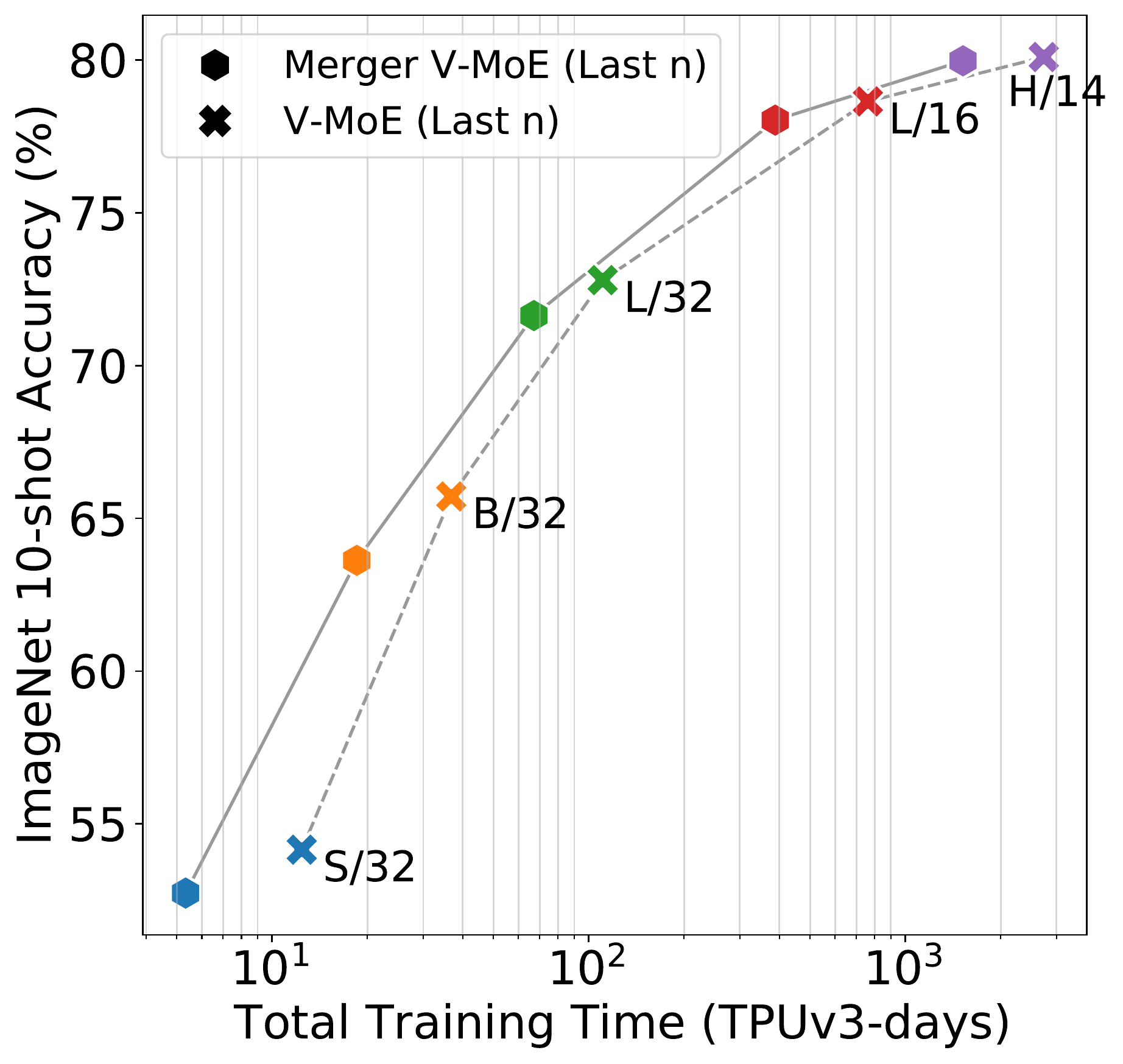}
  \caption{10-shot accuracy on ImageNet.}
  \label{fig:vmoe_upstream_fewshow_runtime_sub2}
\end{subfigure}
\caption{
\textbf{Patch Merger for V-MoE}. Performance versus total training runtime. Colors represent different V-MoE variants, markers represent either standard $\vmoesym{}$ V-MoE or $\varhexagonblack$ Merger V-MoE.
Lines show the Pareto frontier of Merger V-MoE (solid) and V-MoE (dashed).
Figure~\ref{fig:vmoe_upstream_fewshow} in the main text shows performance versus FLOPs for these models.
\label{fig:vmoe_upstream_fewshow_runtime}}
\end{figure}

\begin{figure}
\centering
\begin{subfigure}{.45\textwidth}
  \centering
  \includegraphics[width=\linewidth]{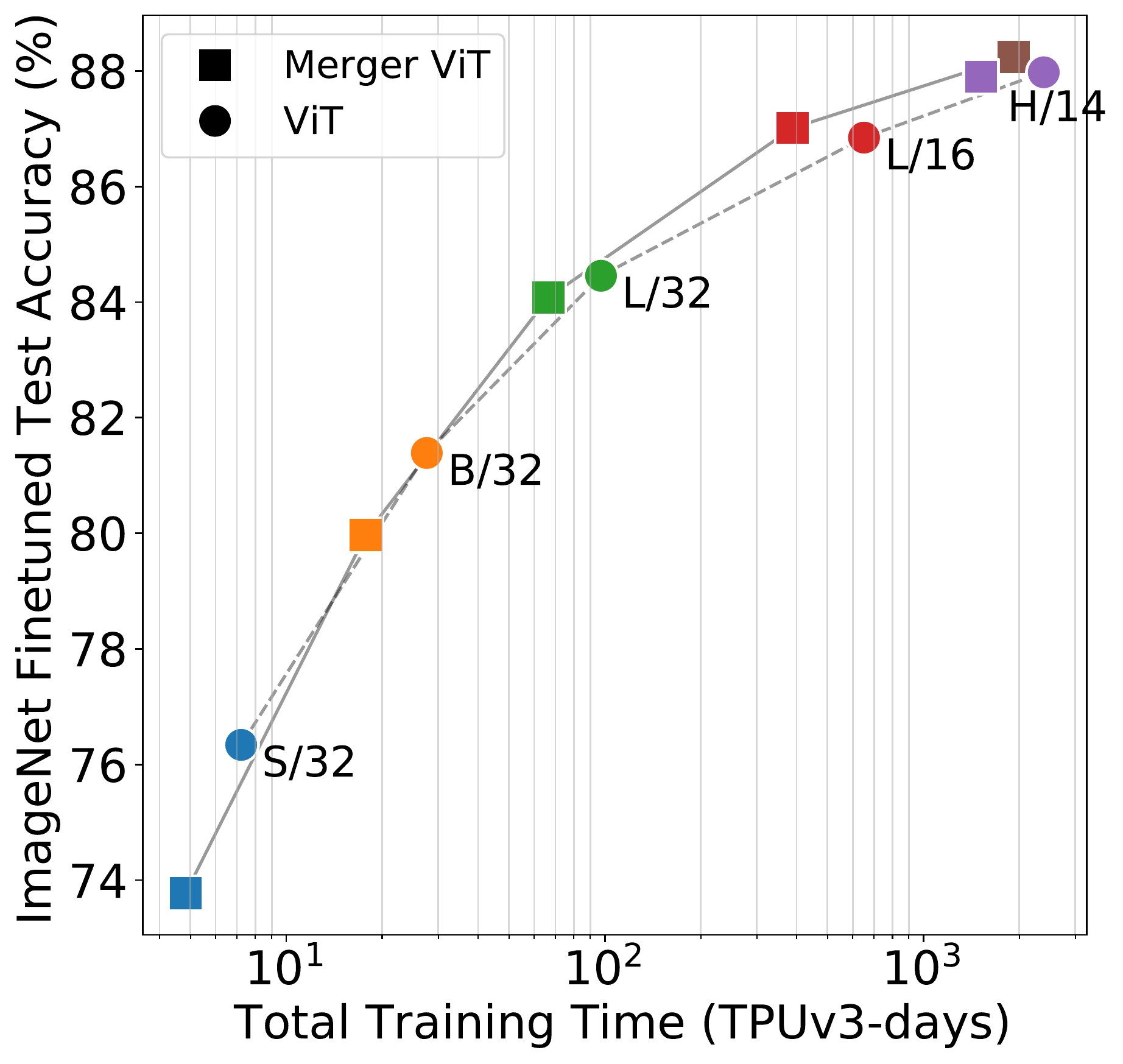}
  \caption{\textbf{ViT.} Finetuning accuracy on Imagenet.}
  \label{fig:downstream_runtime_sub1}
\end{subfigure}%
\begin{subfigure}{.45\textwidth}
  \centering
  \includegraphics[width=\linewidth]{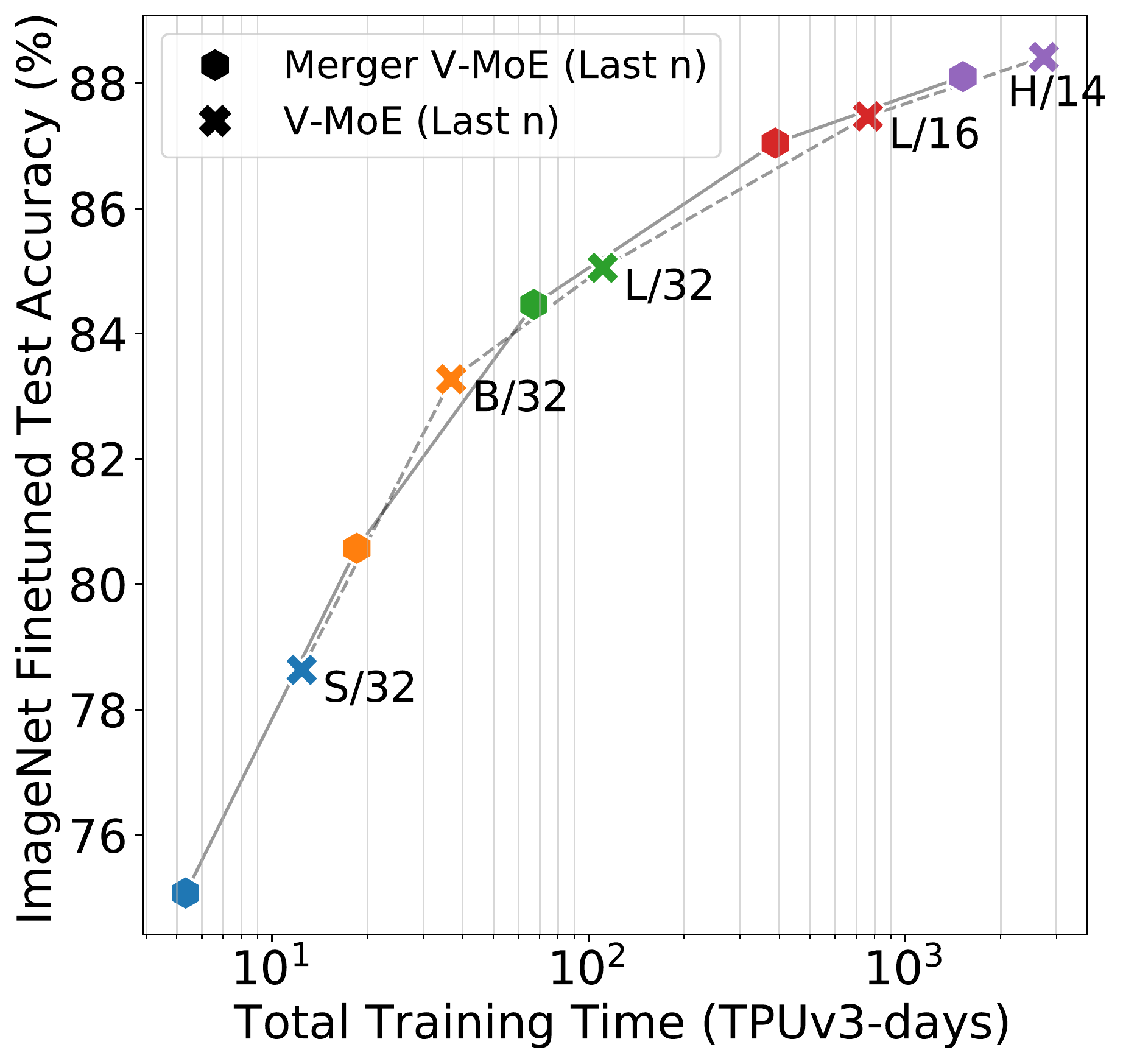}
  \caption{\textbf{V-MoE.} Finetuning accuracy on Imagenet.}
  \label{fig:downstream_runtime_sub2}
\end{subfigure}
\caption{
\textbf{Full finetuning on ImageNet}. Performance versus total training runtime. Colors represent different ViT and V-MoE variants (including H/11 for Merger ViT), markers represent either standard $\densesym{}$ ViT, $\everysym{}$ Merger ViT, $\vmoesym{}$ V-MoE or $\varhexagonblack$ Merger V-MoE.
Lines show the Pareto frontiers in each case.
Figure~\ref{fig:downstream} in the main text shows performance versus FLOPs for these models.
\label{fig:downstream_runtime}}
\end{figure}

\end{document}